# Unethical Research: How to Create a Malevolent Artificial Intelligence

Federico Pistono
Independent Researcher
research@federicopistono.org

Roman V. Yampolskiy
University of Louisville
roman.yampolskiy@louisville.edu

## Abstract

Cybersecurity research involves publishing papers about malicious exploits as much as publishing information on how to design tools to protect cyber-infrastructure. It is this information exchange between ethical hackers and security experts, which results in a well-balanced cyber-ecosystem. In the blooming domain of AI Safety Engineering, hundreds of papers have been published on different proposals geared at the creation of a safe machine, yet nothing, to our knowledge, has been published on how to design a malevolent machine. Availability of such information would be of great value particularly to computer scientists, mathematicians, and others who have an interest in AI safety, and who are attempting to avoid the spontaneous emergence or the deliberate creation of a dangerous AI, which can negatively affect human activities and in the worst case cause the complete obliteration of the human species. This paper provides some general guidelines for the creation of a Malevolent Artificial Intelligence (MAI).

## 1 Introduction

A significant number of papers and books have been published in recent years on the topic of Artificial Intelligence safety and security, particularly with respect to superhuman intelligence [1-13]. Most such publications address unintended consequences of poor design decisions, incorrectly selected ethical frameworks or limitations of systems, which do not share human values and human common sense in interpreting their goals. This paper does not focus on unintentional problems, which might arise as the result of construction of intelligent or superintelligent machines, but rather looks at intentional malice in design. Bugs in code, unrepresentative data, mistakes in design and software poorly protected from black hat hackers can all potentially lead to undesirable outcomes. However, intelligent systems constructed to inflict intentional harm could be a much more serious problem for humanity.

### 1.1 Why the paper on this topic?

Following cybersecurity research paradigm, which involves publishing papers about malicious exploits as much as publishing information on how to design tools to protect cyber-infrastructure, we attempt to summarize general guidelines for development of malicious intelligent software. Information exchanges between hackers and security experts are indispensable to sustainment of a well-balanced cyber-ecosystem in the domain of security. In the flourishing domain of AI safety hundreds of papers have been published on different proposals geared at creation of a safe machine [14], yet nothing, to our knowledge, has been published on how to design a malevolent intelligent system. Availability of such information would be of great value particularly to computer scientists and security experts who have an interest in AI safety, and who are attempting to avoid both the spontaneous emergence and the deliberate creation of a malevolent AI. However, to avoid increasing MAI risks, we avoid publishing any design/pathway details that would be useful to malefactors, not obvious, and not quickly fixable.

### 1.2 Hazardous Intelligent Software

"Computer software is directly or indirectly responsible for controlling many important aspects of our lives. Wall Street trading, nuclear power plants, Social Security compensation, credit histories, and traffic lights are all software controlled and are only one serious design flaw away from creating disastrous consequences for millions of people. The situation is even more dangerous with software specifically designed for malicious purposes, such as viruses, spyware, Trojan horses, worms, and other hazardous software (HS). HS is capable of direct harm as well as sabotage of legitimate computer software employed in critical systems. If HS is ever given the capabilities of truly artificially intelligent systems (e.g., artificially intelligent virus), the consequences unquestionably would be disastrous. Such Hazardous Intelligent Software (HIS) would pose risks currently unseen in malware with subhuman intelligence." [15]. Nick Bostrom, in his typology of information hazards, has proposed the term *artificial intelligence hazard*, which he defines as [16] "computer-related risks in which the threat would derive primarily from the cognitive sophistication of the program rather than the specific properties of any actuators to which the system initially has access."

Addressing specifically superintelligent systems, we can also look at the definition of Friendly Artificial Intelligence (FAI) proposed by Yudkowsky [1] and from it derive a complimentary definition for Unfriendly Artificial Intelligence: a hypothetical Artificial General Intelligence (AGI) that would have a negative rather than positive effect on humanity. Such a system would be capable of causing great harm to all living entities and its values and goals would be misaligned with those of humanity. The system does not have to be explicitly antagonistic to humanity, it is sufficient for it to be neutral to our needs.

An intelligent system could become malevolent in a number of ways, which we can classify into: unintentional and intentional on the part of the designer. Unintentional pathways are most frequently a result of a mistake in design, programming, goal assignment or a result of environmental factors such as failure of hardware. Just like computer viruses and other malware is intentionally produced today, in the future we will see pre-mediated production of hazardous and unfriendly intelligent systems [17]. We are already getting glimpses of such technology in today's research with recently publicized examples involving lying robots [18, 19], black market trading systems [20] and swearing computers [21].

**1.3 Who might be interested in creating Malevolent AI?**
Purposeful creation of MAI can be attempted by a number of diverse agents with varying degrees of competence and success. Each such agent would bring its own goals/resources into the equation, but what is important to understand here is just how prevalent such attempts will be and how numerous such agents can be. Below is a short list of representative entities, it is very far from being comprehensive.
- **Military** developing cyber-weapons and robot soldiers to achieve dominance.
- **Governments** attempting to use AI to establish hegemony, control people, or take down other governments.
- **Corporations** trying to achieve monopoly, destroying the competition through illegal means.
- **Villains** trying to take over the world and using AI as a dominance tool.
- **Black Hats** attempting to steal information, resources or destroy cyberinfrastructure targets.
- **Doomsday Cults** attempting to bring the end of the world by any means.
- **Depressed** looking to commit suicide by AI.
- **Psychopaths** trying to add their name to history books in any way possible.
- **Criminals** attempting to develop proxy systems to avoid risk and responsibility.
- **AI Risk Deniers** attempting to demonstrate that AI is not a risk factor and so ignoring caution.
- **AI Safety Researchers**, if unethical, might attempt to justify funding and secure jobs by purposefully developing problematic AI.

## 2 How to create a Malevolent AI
The literature on AI risk [14] suggests a number of safety measures which should be implemented in any advanced AI project to minimize potential negative impact. Simply inverting the advice would in many cases lead to a purposefully dangerous system. In fact, the number of specific ways in which a malevolent designer may implement hazardous intelligent software is limited only by one's imagination, and it would be impossible to exhaustively review. In "AGI Failures Modes and Levels", Turchin [22] describes a number of ways in which an intelligent system may be dangerous at different stages in its development. Among his examples, AI:
- Hacks as many computers as possible to gain more calculating power
- Creates its own robotic infrastructure by the means of bioengineering
- Prevents other AI projects from finishing by hacking or diversions
- Has goals which include causing suffering
- Interprets commands literally
- Overvalues marginal probability events [22]

Despite impossibility of providing a comprehensive list of failures, we can suggest some general principles, which most experts would agree, are likely to produce MAI by ill-informed but not purposefully malevolent software designers. Those include: Immediate deployment of the system to public networks such as Internet, without testing; Providing system with access to unlimited information including personal information of people, for example massive social networks like Facebook; Giving system specific goals which are not vetted with respect to consequence and unintended side-effects; Putting the system in charge of critical infrastructure such as communication, energy plants, nuclear weapons, financial markets. In the following sections, we provide details on two particular and related types of design mistakes: absence of oversight by qualified experts and secretive nature of project.

**2.1 No oversight boards**
Certain types of research, such as human cloning and the development of biochemical weapons, have been banned or severely restricted due to their unethical nature with respect to laws and human values (as defined by the universal declaration of human rights), and therefore might cause irreparable harm to humanity. Yampolskiy [23] proposed to consider Artificial General Intelligence (AGI) research unethical, as "*true AGIs will be capable of universal problem solving and recursive self-improvement. Consequently they have potential of outcompeting humans in any domain essentially making humankind unnecessary and so subject to extinction*" and "*may possess a type of consciousness comparable to the human type making robot suffering a real possibility and any experiments with AGI unethical for that reason as well.*" The creation of an oversight board, a team of experts in artificial intelligence that evaluates each research proposal and decides if the proposal is for standard AI or may lead to the development of a full blown AGI, would either stop (albeit very unlikely), severely limit, or at the very least slow down the process of creation of an uncontrolled AGI.

If a group decided to create a MAI, it follows that preventing a global oversight board committee from coming to existence would increase its probability of succeeding. This can be accomplished in a number of ways, most of which exploit one or more of the innumerable cognitive biases humans are victims of.

The first strategy is to disseminate conflicting information that would create doubt in the public's imagination about the dangers and opportunities of AGI research. One lesson can be learned from the public "debate" around the theory of evolution through means of natural selection and that of anthropogenic climate change. Even though the scientific consensus among experts is overwhelming on both scientific theories, the public perception is that there is an ongoing debate in the scientific community and that the matter is not settled in either case. This has been accomplished by means of politicization of the issue, or by associating a religious ideology to either side. Similarly, an individual or group with the intention of creating a MAI would attempt to increase polarization in the debate, either through lobbying, marketing, by disseminating false information, by publishing polarizing news stories on popular blogs and anti-establishment websites, by creating conspiratorial hypothesis that feed the public's imagination and the anti-government sentiment, or by attracting the attention of political parties or influential religious groups, just to name a few. The confusion generated would create an environment nebulous enough to delay or even prevent the creation of the global oversight board.

A second option is to convince the government of a country, preferably a small one or a micro-state, not to sign the AGI-prevention treaty that all other major nations would eventually subscribe to. Again, it is not difficult to imagine how this could be achieved. The simplest way would be to simply bribe government officials into accepting this condition. Other means include promises of greatness as a result of being the first country to have access to the first AGI ever created, to intimidation, extortion, or even a simple leverage of the ignorance that they might have on the subject of AGI dangers. In all cases, only one condition needs to be satisfied in order for the plan to succeed. Given the vast heterogeneity of human responses and the long list of cognitive biases one can leverage, it is reasonable to assume that at least one exception will be found and exploited, by someone at some point.

**2.2 Closed source code**
One of the key pieces in both the purposeful and the unintended creation of malevolent artificial intelligence is the ability to run on closed-source, proprietary software and hardware. It is well known among cryptography and computer security experts that closed-source software and algorithms are less secure than their Free and Open Source (FOS) counterpart, they are more prone to vulnerabilities, exploits, deliberate or accidental malevolent use

[24]. While FOS is not a sufficient condition for security, Kerckhoffs' principle shows that closed-source software and security through obscurity are bad practices [25]. It is also known that both closed source software and hardware have been manipulated and tampered with by intelligence agencies and institutions, whether with or without the knowledge or consent of the companies creating the specialized hardware or programming the software running on it [26-28]. In this environment, any group with the intent of creating a malevolent artificial intelligence would find the ideal conditions for operating in quasi-total obscurity, without any oversight board and without being screened or monitored, all the while being protected by copyright law, patent law, industrial secret, or in the name of "national security".

Furthermore, a much broader case deals with the unintended consequences of running sophisticated and powerful non-free software and hardware, even without malicious intents. It has been shown that the indiscriminate collection of personal data and information about people is prone to unintended uses that can backfire, go out of control and have ramifications outside its original intent. By deliberately compromising the security of a system, either hardware or software, the integrity of the system's operations is put at risk and it is at the mercy of anyone capable of running the exploit that was created (i.e. to collect people's personal information). Any system having a universal backdoor (i.e. anything that implements Digital Rights Management (DRM) restrictions) can be used for virtually any purpose, without letting the user know that the system has been compromised or that it's being used for unintended purposes, whatever they might be.

The very existence of non-free software and hardware puts humanity at a greater risk, it can foster or accelerate the process of the creation of an artificial entity that can outcompete or control humans in any domain, making humankind unnecessary, controllable, or even subject to extinction, and should therefore be considered unethical. The authors recognize the likelihood of non-free software be rendered illegal or banned globally is extremely low, therefore greatly increasing the probability of the creation of a MAI and that of human subjugation or extinction. At the same time, wide distribution of an open-source code for AI may make it possible for a number of additional agents, which otherwise lack technical sophistication, to engage in conversion of a safe AI product into a malevolent one. It is unclear what the solution to prevent this would be. It is arguable that greater access to the source code by a greater number of people would give more chances to prevent or fix problems, given that historically there is a greater number of individuals and organization whose goal is to fix bugs and keep systems safe rather than creating damage. On the other hand, terrorists and criminals often have some success in identifying and exploiting new vulnerabilities despite great defensive efforts.

Recent announcement of OpenAI, a non-profit artificial intelligence research company whose goal is to "advance digital intelligence in the way that is most likely to benefit humanity as a whole, unconstrained by a need to generate financial return", with over $1 billion pledged for its development, is a promising step in the right direction. Particularly, if the research will also be focused on monitoring and preventing early MAI attacks, although it's too early to say at this point.

# 3 Societal Impact

In this section, we will review potential impact of MAI on economy, government, legislation and military.

### 3.1 Economy/Unemployment

The dangers of technological unemployment, which states that technology eventually advances so rapidly that it creates a non-cyclical, structural unemployment problem in the labor market, are starting to be considered as a serious matter in academia. Research by Pistono [29], Frey and Osborne [30], suggests that we might already see early sign of this process today and that as much 47% of all jobs in the U.S. could be automated. Further analysis by Bowels [31] indicates that the same trend would apply to Europe as well, with numbers fluctuating between 61.95% in Romania and 47.17% in the U.K, averaging 54% across the EU-28. The proposed solutions, which range from tax reforms, to promoting entrepreneurship and adopting an Unconditional Basic Income [32], are non-homogeneous, polarized, and not universally accepted.

The topic is still in its infancy, and the prospect of technological unemployment has not been widely accepted by most governments, which struggle to recognize its very existence, let alone come to an agreement on a coherent solution that would be practical across the globe. Given the historical context and precedents, we expect governments to react non-homogeneously and at different times. This creates the ideal fertile ground for a MAI to take over without encountering a strong, coherent, and organized opposition. By exploiting the natural tendency of companies to want to increase their productivity and profits, a MAI could lend its services to corporations, which would be more competitive than most, if not all humans workers, making it desirable for a company to employ the quasi-zero marginal cost AI, in place of the previously employed expensive and inefficient human labor.

A MAI could also, reasonably, run on closed-source software, and operate on the cloud under SaaS subscriptions (Software As A Service**,** a software licensing and delivery model in which software is licensed on a subscription basis and is centrally hosted), or by licensing a subset of its functions to install on the companies' servers, where it could run small exploit programs and worms. These programs would be very hard to detect, and could take over servers and global operations very quickly, and execute their goals accordingly. Even under the assumption that a MAI would not succeed in deploying such exploits, viruses, and worms, there is still a high likelihood that it would become the prevalent provider of services across the world, effectively giving it the ability to control global food markets, power stations, transportation, financial markets, logistics and operations, just to name a few industries.

A particularly sophisticated MAI would be able to predict which behaviors would catch the attention of external observers who look for signs of its emergence, and would mimic the functioning of a normal company or even an ecosystem of companies that operate at the global level, including employing human beings, so as not to draw suspicion. The structure of the corporation theoretically allows for such hierarchical systems to function, where workers don't need to know who or what is controlling the whole, and where work is compartmentalized appropriately. Additionally, corporate personhood in the U.S. recognizes a corporation as an individual in the eyes of the law, the same applying also to the City of London in the United Kingdom. With this in place, a MAI could conceivably create a network of corporations that can dominate the market within the span of a decade or less, where the creation of structural, irreversible unemployment of more than half of all jobs would be the least worrisome outcome.

### 3.2 Government

Governments play a crucial role in the functioning of all modern societies. The balance of powers and oversight has been the subject of controversy and heated debate since the first primitive forms of organized societal structures, which over time evolved into modern forms of governments.

*Coups d'état*, the sudden and illegal seizure of a state either by force or through corruption, intimidation, or economic hitmen, can create instability, uprisings, civil wars, poverty and in the worst cases a total collapse of the country. Throughout history, humans have successfully staged *Coups d'état* by disseminating false information, staging assassinations, and other relatively simple operations that could be easily accomplished by a MAI. In addition, a sufficiently advanced MAI with access to the global system of communication, internet and social media, can stage multiple *Coups* simultaneously, run simulations of possible ramification of each attack, incite governments and populations against one another, and feed exponentially a planetary chaos machine to bring the maximum level of instability and therefore compromising the safety and ability to respond to further attacks, which can be combined with judicial, legislative, and military action. The safest course of action for a MAI would probably be to twist the knobs just enough to create global chaos and let humanity destroy or weaken itself to the point of being extremely vulnerable, while remaining undetected and unsuspected, and thus maintaining its existence with minimal risk. If governments believe they are fighting terrorist groups, either internal or foreign, and are constantly in a state of paranoia, they will continue to distrust one another, engaging in cyber, economic, and even physical attacks, furthering the goal of the MAI.

All this can be augmented and tweaked to perfection by scanning through the data available on social media, emails, text messages, phone calls and other means of electronic communication. The more the data is available and unencrypted, the more a MAI can understand how to best respond and stage the next wave of attacks. The NSA, by illegally installing backdoors into hardware and software worldwide, collecting personal information and metadata of all internet users that go through US-controlled servers, has weakened the security of our communication systems, and has made us more vulnerable to espionage and terrorist attacks [33-35]. Security agencies in other countries have also been involved in similar activities, making the problem worse. A MAI can capitalize on the fundamental insecurity of these systems, and have access to an unprecedented level of information regarding all human activities, and therefore have almost perfect information on which to simulate possibilities, stage attacks and diversions to reach its goals while remaining undetected and unsuspected.

## 3.3 Legislative

Any action or strategy deployed by a MAI at the expense of humans and their life-supporting systems can be amplified and accelerated by legislation in support of such plans. Research by Lessig [36] shows that the extreme concentration of power, influence, and decision-making in the hands of a few has made it possible to completely decouple the will of the people from the results of legislations of countries such as the United States, where 0.05% of the population can consistently pass laws that are against the will of the majority. It follows that a MAI would seek control of the House, the Senate, and other bodies through lobbying, at a fraction of the energy and time expenditure (cost) that would require via other means. Hidden and obscure pro-MAI activities legislation can feed into a positive feedback cycle that further increases the influence of a MAI and the likelihood of developing its goals. The more power becomes centralized and influence is in the hands of few, the easier it will be for a MAI to seek control of the world. It is also very likely that MAI itself would attempt to enter politics [37] or penetrate our judicial system [38] to avoid having to rely on human assistants.

## 3.4 Military

In the authors' view, military action and weapons against humans, whilst vivid in the public's imagination and in sci-fi stories, are very unlikely to be used by a MAI, unless all other options fail. Armed forces and military interventions are noticeable and very hard to hide. As explained earlier, a MAI will make use of all possible means to go unnoticed and unsuspected for the longest period possible, so as to not give time for humans to assemble and strategize counter-attacks or effective defenses. The direct use of force when other more efficient options are available is an irrational strategy, of which humans are subject to because of their evolutionary biological baggage, but a MAI would have no reason for it, and therefore would delay or even avoid its use altogether.

Should necessity arise, striking a global conflict will be a very easy task for a MAI. The total number of nuclear weapons is estimated at 10,144 [39], many of which are stored in aging facilities, without the proper safety standards, underfunded, and numerous scandals have put into question the ability of operators to keep them safe [40-43]. If trained officers and even civilians have been able to breach the security of these systems, any sufficiently advanced MAI can accomplish the goal with ease. As mentioned before, the path of least resistance is the most logical and efficient course of action. Rather than announcing and launching a deliberate attack against humanity, it would be much easier to simply exploit the security flaws of existing nuclear storage facilities and their human personnel. Of course, the possibility remains that MAI will take over military robots, drones and cyber weapons to engage in a direct attack by exploiting autonomous capabilities of such weapons [44]. If the goal were to exterminate humanity, one of the easiest paths would be to design an extremely deadly virus or bacteria, which can stay dormant and undetected until it reaches the entirety the world's population, and is then activated wirelessly via nanobots simultaneously and ubiquitously, killing all humans in a matter of minutes or hours.

## 4. Conclusions

A lot has been written about problems which might arise as a result of arrival of true AI, either as a direct impact of such invention or because of a programmer's error [14, 45-52]. Intentional malice in design has not been addressed in scientific literature, but it is fair to say that, when it comes to dangers from a purposefully unethical superintelligence, technological unemployment may be the least of our concerns. According to Bostrom's orthogonality thesis an AI system can have any combination of intelligence and goals [53]. Such goals can be introduced either via initial design or in case of an "off the shelf" superintelligence introduced later – "just add your own goals". Consequently, depending on whose bidding the system is doing [governments, corporations, sociopaths, dictators, military industrial complexes, doomsdayers, etc.] it may attempt to inflict damage unprecedented in the history of humankind or perhaps inspired by it. The world is not short of terrible people and, as it is famously stated, "*absolute power corrupts absolutely*". Even in absence of universally malicious motives [the destruction of the human species], the simple will to subjugate or to control certain groups of people or to gain advantage over someone else is enough to make a MAI desirable for some entity or organization. In the order of (subjective) undesirability from least damaging to ultimately destructing, a malevolent superintelligence may attempt to:

- Takeover (implicit or explicit) of resources such as money, land, water, rare elements, organic matter, internet, computer hardware, etc. and establish monopoly over access to them;

- Take over political control of local and federal governments as well as of international corporations, professional societies, and charitable organizations;
- Reveal informational hazards [16];
- Set up a total surveillance state (or exploit an existing one), reducing any notion of privacy to zero including privacy of thought;
- Force merger (cyborgization) by requiring that all people have a brain implant which allows for direct mind control/override by the superintelligence;
- Enslave humankind, meaning restricting our freedom to move or otherwise choose what to do with our bodies and minds. This can be accomplished through forced cryonics or concentration camps;
- Abuse and torture humankind with perfect insight into our physiology to maximize amount of physical or emotional pain, perhaps combining it with a simulated model of us to make the process infinitely long;
- Commit specicide against humankind, arguably the worst option for humans as it can't be undone;
- Destroy/irreversibly change the planet, a significant portion of the Solar system, or even the universe;
- *Unknown Unknowns.* Given that a superintelligence is capable of inventing dangers we are not capable of predicting, there is room for something much worse but which at this time has not been invented.

*Artificial Intelligence Safety Engineering* (AISE) is an emerging interdisciplinary field of science concerned with making current and future Artificially Intelligent systems safe and secure. Engineers, Computer Scientists, Philosophers, Economists, Cognitive Scientists and experts in many other fields are starting to realize that despite differences in terminology and approaches they are all working on a common set of problems associated with control over independent and potentially malevolent intelligent agents. Currently such research is conducted under such diverse labels as: AI ethics, Friendly AI, Singularity Research, Superintelligence Control, Software Verification, Machine Ethics, and many others. This leads to suboptimal information sharing among practitioners, confusing lack of common terminology and as a result limited recognition, publishing and funding opportunities. It is our hope that by outlining main dangers associated with future intelligent systems we are able to inspire a more cohesive research program aimed at counteracting intentional development of dangerous software. Perhaps, even a malevolent AI could be repaired and turned into a Friendly AI with a sufficient level of understanding on the part of AI safety researchers. Of course, those attempting to purposefully introduce MAI into the world will no doubt provide their systems with a sophisticated self-defense module preventing intervention by AI safety researchers attempting to neutralize it.

Finally, we are issuing a challenge to the AI Safety community to continue this work by discovering and reporting specific problems, which may lead to MAI along with request that publication of concrete solutions to such problems take place prior to open publication of the problematic source code.

## Acknowledgements

Roman Yampolskiy expresses appreciation to Elon Musk and FLI for partially funding his work via project grant: "Evaluation of Safe Development Pathways for Artificial Superintelligence" awarded to Global Catastrophic Risk Institute, Seth Baum (PI). The authors are grateful to Yana Feygin, Seth Baum, James Babcock, János Kramár and Tony Barrett for valuable feedback on an early draft of this paper. Views in this paper are those of the authors, and do not necessarily represent the views of FLI, GCRI, or others.## References

1. Yudkowsky, E., *Artificial Intelligence as a Positive and Negative Factor in Global Risk*, in *Global Catastrophic Risks*, N. Bostrom and M.M. Cirkovic, Editors. 2008, Oxford University Press: Oxford, UK. p. 308-345.
2. Chalmers, D., *The Singularity: A Philosophical Analysis.* Journal of Consciousness Studies, 2010. **17**: p. 7-65.
3. Bostrom, N., *Ethical Issues in Advanced Artificial Intelligence.* Review of Contemporary Philosophy, 2006. **5**: p. 66-73.
4. Yampolskiy, R.V., *Utility Function Security in Artificially Intelligent Agents.* Journal of Experimental and Theoretical Artificial Intelligence (JETAI), 2014: p. 1-17.
5. Armstrong, S., A. Sandberg, and N. Bostrom, *Thinking inside the box: Controlling and using an oracle ai.* Minds and Machines, 2012. **22**(4): p. 299-324.

6. Muehlhauser, L. and L. Helm, *The Singularity and Machine Ethics*, in *In The Singularity Hypothesis: A Scientific and Philosophical Assessment, edited by A. Eden, J. Søraker, J. Moor, and E. Steinhart*. 2012, Springer: Berlin, Heidelberg.
7. Hibbard, B., *Super-Intelligent Machines.* Computer Graphics, 2001. **35(1)**: p. 11-13.
8. Loosemore, R. and B. Goertzel, *Why an intelligence explosion is probable*, in *Singularity Hypotheses*. 2012, Springer. p. 83-98.
9. Omohundro, S.M., *The Basic AI Drives*, in *Proceedings of the First AGI Conference, Volume 171, Frontiers in Artificial Intelligence and Applications, P. Wang, B. Goertzel, and S. Franklin (eds.)*. February 2008, IOS Press.
10. Yampolskiy, R.V., *The Space of Possible Mind Designs*, in *Artificial General Intelligence*. 2015, Springer. p. 218-227.
11. Yampolskiy, R.V., *Artificial Superintelligence: a Futuristic Approach*. 2015: Chapman and Hall/CRC.
12. Yampolskiy, R.V., *On the Limits of Recursively Self-Improving AGI*, in *The Eighth Conference on Artificial General Intelligence*. July 22-25, 2015: Berlin, Germany.
13. Yampolskiy, R.V., *Analysis of Types of Self-Improving Software*, in *The Eighth Conference on Artificial General Intelligence*. July 22-25, 2015: Berlin, Germany.
14. Sotala, K. and R.V. Yampolskiy, *Responses to catastrophic AGI risk: A survey.* Physica Scripta January 2015. **90**.
15. Yampolskiy, R.V., *Leakproofing Singularity - Artificial Intelligence Confinement Problem.* Journal of Consciousness Studies (JCS), 2012. **19(1-2)**: p. 194–214.
16. Bostrom, N., *Information Hazards: A Typology of Potential Harms From Knowledge.* Review of Contemporary Philosophy, 2011. **10**: p. 44-79.
17. Yampolskiy, R.V., *Taxonomy of Pathways to Dangerous AI*, in *30th AAAI Conference on Artificial Intelligence (AAAI-2016). 2nd International Workshop on AI, Ethics and Society (AIEthicsSociety2016)*. February 12-13th, 2016: Phoenix, Arizona, USA.
18. Castelfranchi, C., *Artificial liars: Why computers will (necessarily) deceive us and each other.* Ethics and Information Technology, 2000. **2**(2): p. 113-119.
19. Clark, M.H., *Cognitive illusions and the lying machine: a blueprint for sophistic mendacity*. 2010, Rensselaer Polytechnic Institute.
20. Cush, A., *Swiss Authorities Arrest Bot for Buying Drugs and Fake Passport*, in *Gawker*. January 22, 2015: http://internet.gawker.com/swiss-authorities-arrest-bot-for-buying-drugs-and-a-fak-1681098991.
21. Smith, D., *IBM's Watson Gets A 'Swear Filter' After Learning The Urban Dictionary*, in *International Business Times*. January 10, 2013: http://www.ibtimes.com/ibms-watson-gets-swear-filter-after-learning-urban-dictionary-1007734.
22. Turchin, A., *A Map: AGI Failures Modes and Levels*, in *LessWrong*. July 10 2015: http://lesswrong.com/lw/mgf/a_map_agi_failures_modes_and_levels/.
23. Yampolskiy, R.V., *Artificial intelligence safety engineering: Why machine ethics is a wrong approach*, in *Philosophy and Theory of Artificial Intelligence*. 2013, Springer Berlin Heidelberg. p. 389-396.
24. Schneier, B., *Open Source and Security*. 1999: https://www.schneier.com/crypto-gram/archives/1999/0915.html#OpenSourceandSecurity.
25. Hoepman, J.-H. and B. Jacobs, *Increased security through open source.* Communications of the ACM, 2007. **50**(1): p. 79-83.
26. Appelbaum, J., J. Horchert, and C. Stöcker, *Shopping for Spy Gear: Catalog Advertises NSA Toolbox*, in *Der Spiegel*. 2013: http://www.spiegel.de/international/world/catalog-reveals-nsa-has-back-doors-for-numerous-devices-a-940994.html.
27. Schneier, B., *CROSSBEAM: NSA Exploit of the Day*. 2014: https://www.schneier.com/blog/archives/2014/02/crossbeam_nsa_e.html.
28. Schneier, B., *More about the NSA's Tailored Access Operations Unit*. 2013: https://www.schneier.com/blog/archives/2013/12/more_about_the.html.
29. Pistono, F., *Robots will steal your job, but that's ok: how to survive the economic collapse and be happy*. 2012: CreateSpace.

30. Frey, C.B. and M.A. Osborne, *The future of employment: how susceptible are jobs to computerisation?* Sept, 2013. **17**: p. 2013.
31. Bowels, J., *The computerisation of European jobs*. 2014: http://bruegel.org/2014/07/the-computerisation-of-european-jobs/.
32. Haagh, A. and M. Howard, *Basic Income Studies*. 2012. **7(1)**.
33. Peha, J.M., *The Dangerous Policy of Weakening Security to Facilitate Surveillance.* Available at SSRN: http://ssrn.com/abstract=2350929. http://dx.doi.org/10.2139/ssrn.2350929, October 4, 2013.
34. Paterson, K., et al., *Open Letter From UK Security Researchers*, in *University of Bristol*. 2013: http://bristolcrypto.blogspot.co.uk/2013/09/open-letter-from-uk-security-researchers.html.
35. Abadi, M. and e. al., *An Open Letter from US Researchers in Cryptography and Information Security*. January 24, 2014: http://masssurveillance.info.
36. Lessig, L., *Republic, Lost: How Money Corrupts Congress--and a Plan to Stop It*. 2011, New York, NY: Twelve.
37. Pellissier, H., *Should Politicians be Replaced by Artificial Intelligence?*, in *IEET*. June 12, 2015: http://ieet.org/index.php/IEET/more/pellissier20150612.
38. Dunn, T., *Exploring the possibility of using intelligent computers to judge criminal offenders*. 1995: http://www.fdle.state.fl.us/Content/getdoc/814750b4-4546-49f0-9be2-90fe772257f7/Dunn.aspx.
39. *Nuclear Notebook, Bulletin of the Atomic Scientists*. 2014: http://thebulletin.org.
40. *Security Improvements at the Y-12 National, Security Complex*, in *U.S. Department of Energy, Office of Inspector General, Office of Audits and Inspections*. 2015: http://energy.gov/sites/prod/files/2015/09/f26/DOE-IG-0944.pdf.
41. Smith, R.J., *How an 82-year-old exposed security lapses at nuclear facilities*, in *The Center for Public Integrity*. September 12, 2012: http://www.publicintegrity.org/2012/09/12/10851/how-82-year-old-exposed-security-lapses-nuclear-facilities.
42. *Audit Report Security at the Nevada National Security Site*, in *U.S. Department of Energy, Office of Inspector General, Office of Audits and Inspections* 2015: http://energy.gov/sites/prod/files/2015/05/f22/OAS-L-15-06.pdf.
43. *US nuclear launch officers suspended for 'cheating'*, in *BBC*. January 16, 2014: http://www.bbc.com/news/world-us-canada-25753040
44. Arkin, R.C., *The case for ethical autonomy in unmanned systems.* Journal of Military Ethics, 2010. **9**(4): p. 332-341.
45. Majot, A.M. and R.V. Yampolskiy. *AI safety engineering through introduction of self-reference into felicific calculus via artificial pain and pleasure*. in *Ethics in Science, Technology and Engineering, 2014 IEEE International Symposium on*. 2014. IEEE.
46. Yampolskiy, R. and J. Fox, *Safety Engineering for Artificial General Intelligence.* Topoi, 2012: p. 1-10.
47. Yampolskiy, R.V. and J. Fox, *Artificial General Intelligence and the Human Mental Model.* Singularity Hypotheses: A Scientific and Philosophical Assessment, 2013: p. 129.
48. Yampolskiy, R.V., *What to Do with the Singularity Paradox?*, in *Philosophy and Theory of Artificial Intelligence*. 2013, Springer Berlin Heidelberg. p. 397-413.
49. Yampolskiy, R. and M. Gavrilova, *Artimetrics: Biometrics for Artificial Entities.* IEEE Robotics and Automation Magazine (RAM), 2012. **19**(4): p. 48-58.
50. Yampolskiy, R., et al., *Experiments in Artimetrics: Avatar Face Recognition.* Transactions on Computational Science XVI, 2012: p. 77-94.
51. Ali, N., D. Schaeffer, and R.V. Yampolskiy, *Linguistic Profiling and Behavioral Drift in Chat Bots.* Midwest Artificial Intelligence and Cognitive Science Conference, 2012: p. 27.
52. Gavrilova, M. and R. Yampolskiy, *State-of-the-Art in Robot Authentication [From the Guest Editors].* Robotics & Automation Magazine, IEEE, 2010. **17**(4): p. 23-24.
53. Bostrom, N., *The superintelligent will: Motivation and instrumental rationality in advanced artificial agents.* Minds and Machines, 2012. **22**(2): p. 71-85.